\documentclass[10pt,twocolumn,letterpaper]{article}
\usepackage{colortbl}
\usepackage{multirow}
\usepackage{makecell}
\usepackage{bm}
\usepackage{float}
\usepackage{pifont}
\usepackage{balance}
\usepackage{amsmath, amssymb}
\usepackage{dblfloatfix}
\usepackage[accsupp]{axessibility}  
\usepackage[pagenumbers]{iccv} 

\definecolor{iccvblue}{rgb}{0.21,0.49,0.74}
\usepackage[pagebackref,breaklinks,colorlinks,allcolors=iccvblue]{hyperref}

\def\paperID{} 
\def\confName{ICCV}
\def\confYear{2025}

\title{CorrCLIP: Reconstructing Patch Correlations in CLIP 

for Open-Vocabulary Semantic Segmentation}

\author{
Dengke Zhang$^1$~~~~~~~~~~
Fagui Liu$^{1,2}$\thanks{Corresponding authors.}~~~~~~~~~~
Quan Tang$^2$\footnotemark[1]
\smallskip 
\\
$^1$South China University of Technology
~~~~~~~~~~
$^2$Pengcheng Laboratory
\\
{\tt\small 
csdk@mail.scut.edu.cn,
fgliu@scut.edu.cn,
tangq@pcl.ac.cn}
}

\begin{document}
\maketitle
\begin{abstract}
Open-vocabulary semantic segmentation aims to assign semantic labels to each pixel without being constrained by a predefined set of categories. While Contrastive Language-Image Pre-training (CLIP) excels in zero-shot classification, it struggles to align image patches with category embeddings because of its incoherent patch correlations. This study reveals that inter-class correlations are the main reason for impairing CLIP's segmentation performance. Accordingly, we propose CorrCLIP, which reconstructs the scope and value of patch correlations.
Specifically, CorrCLIP leverages the Segment Anything Model (SAM) to define the scope of patch interactions, reducing inter-class correlations. To mitigate the problem that SAM-generated masks may contain patches belonging to different classes, CorrCLIP incorporates self-supervised models to compute coherent similarity values, suppressing the weight of inter-class correlations. Additionally, we introduce two additional branches to strengthen patch features' spatial details and semantic representation. Finally, we update segmentation maps with SAM-generated masks to improve spatial consistency. 
Based on the improvement across patch correlations, feature representations, and segmentation maps, CorrCLIP achieves superior performance across eight benchmarks.
Codes are available at: \url{https://github.com/zdk258/CorrCLIP}.
\end{abstract}    
\section{Introduction}
\label{sec:intro}
Open-vocabulary semantic segmentation (OVSS) \cite{bucher2019zero,xian2019semantic,zhao2017open} aims to partition an image into multiple segments and assign a corresponding category to each segment based on category descriptions. Contrastive Language-Image Pre-training (CLIP) \cite{clip} model, trained on large-scale image-text pair datasets, shows remarkable zero-shot classification capabilities, providing a viable solution for OVSS.

\begin{figure}
    \setlength{\belowcaptionskip}{-8pt}
  \centering
   \includegraphics[width=1\linewidth]{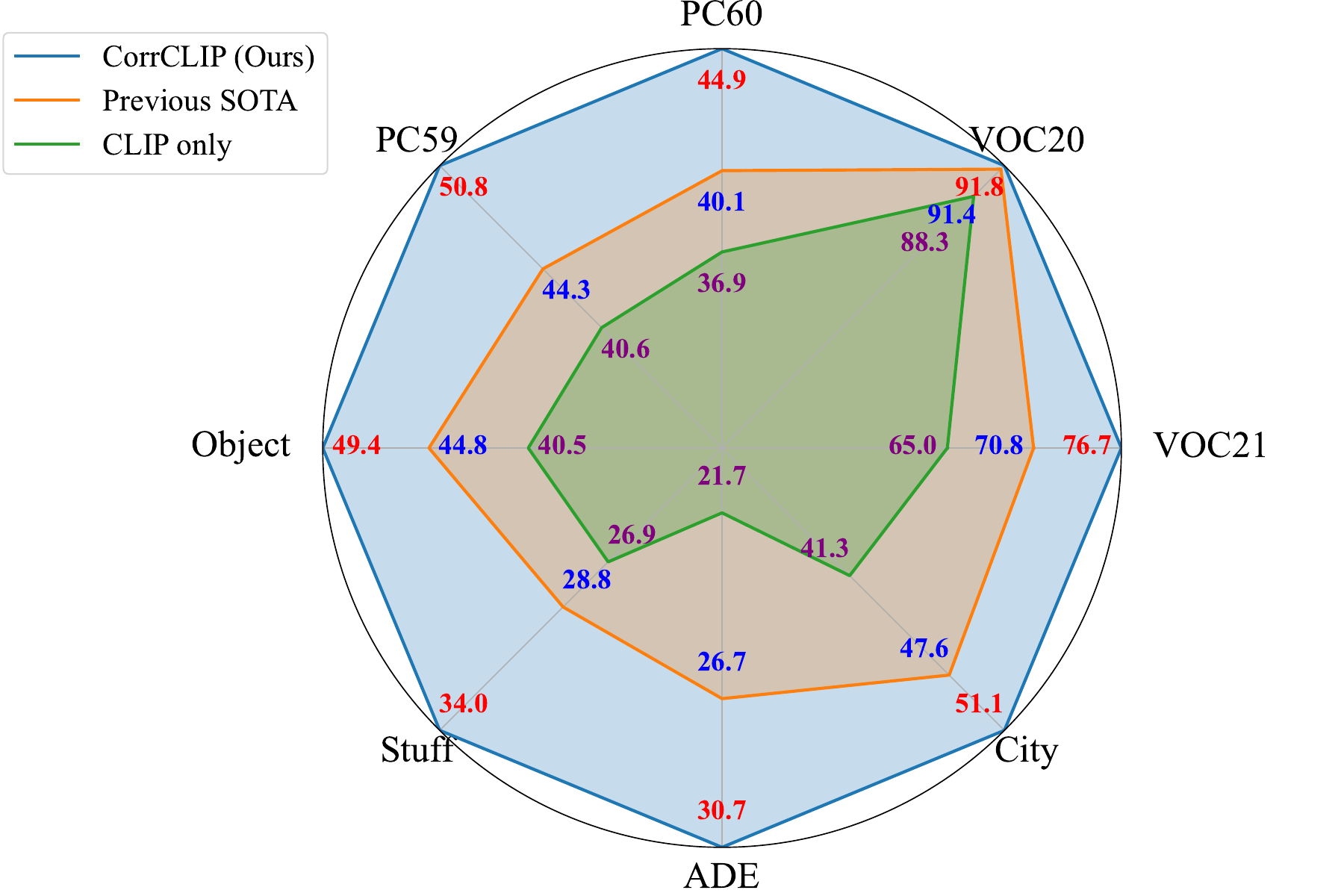}
   \caption{Comparison of OVSS performance on eight benchmarks. ``CLIP only'' refers to the best performance when only using CLIP.}
   \label{radar}
\end{figure}

Applying CLIP to OVSS requires aligning image patches with corresponding category embeddings. However, due to CLIP's contrastive image-text training objective, it prioritizes capturing global semantics, resulting in insufficient discriminability between image patches. ClearCLIP \cite{lan2024clearclip} demonstrates that removing residual connections and feed-forward network from the last layer of Vision Transformer (ViT) \cite{vit} in CLIP’s visual encoder makes patches more distinguishable. Therefore, the key to improving performance lies in the remaining self-attention layer. Self-attention captures correlations between image patches to aggregate contextual information. Although existing methods \cite{clip-srugery, sclip, proxyclip} modify the self-attention to improve CLIP's segmentation performance, they do so without explicitly identifying the specific patch correlations that impair CLIP's segmentation capability, limiting further performance gains. To this end, we reveal that the main impediment is inter-class correlations and propose CorrCLIP to reduce the adverse effect of inter-class correlations. As shown in \cref{radar}, CorrCLIP significantly expands the performance boundaries of OVSS.

\begin{figure}[t]
  \centering
  \setlength{\belowcaptionskip}{-8pt}
  \begin{subfigure}{1\linewidth}
  \setlength{\abovecaptionskip}{0pt} 
  \setlength{\belowcaptionskip}{0pt}
    \includegraphics[width=1\linewidth]{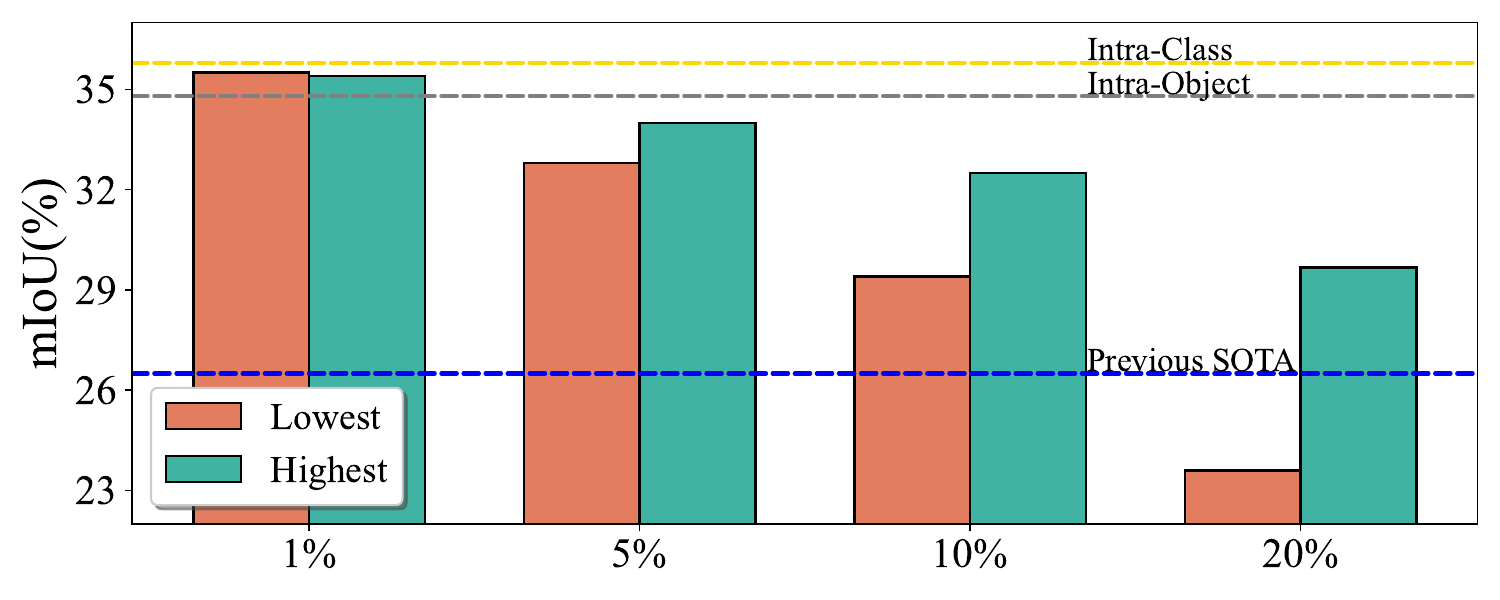}
    \caption{}
    \label{patch1}
  \end{subfigure}
  
  \vspace{-2pt}
  
  \begin{subfigure}{1\linewidth}
    \setlength{\abovecaptionskip}{0pt} 
    \setlength{\belowcaptionskip}{0pt}
    \includegraphics[width=1\linewidth]{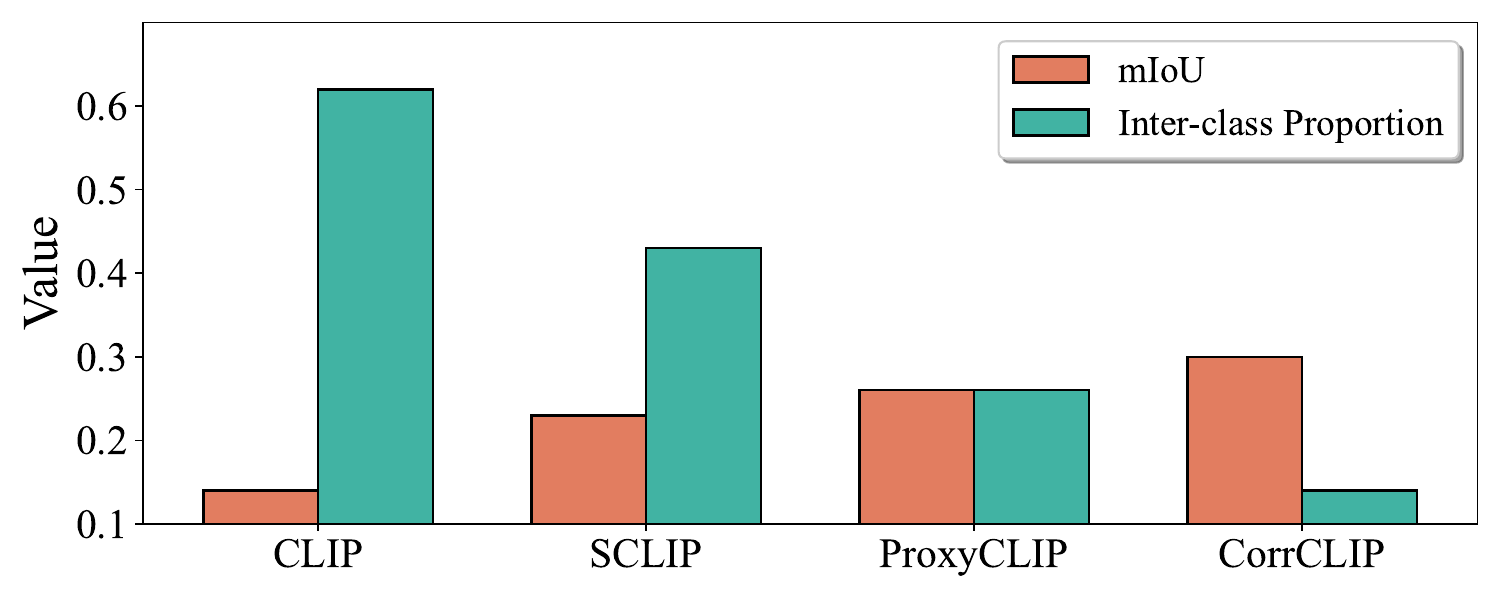}
    \caption{}
    \label{patch2}
  \end{subfigure}
  \caption{Impact of scope of patch correlations on performance of COCO Stuff. (a) The impact of intra-class and intra-object correlations and different proportions of inter-class correlations on segmentation performance. ``Lowest'' and ``Highest'' denote the interactions with the least and most similar inter-class patches, respectively. (b) The inverse relationship between the performance and the proportion of inter-class correlations in current methods.}
  \label{patch}

\end{figure}

The ideal patch correlations of CLIP for the segmentation task should enhance the distinguishability of patches by ensuring patch features from different classes remain distinct while also aggregating semantic information effectively to align patches with their corresponding category vectors. We divide patch correlations into two types: intra-class and inter-class correlations. Intuitively, intra-class correlations can achieve the above two points and enhance CLIP's segmentation capability. However, the impact of inter-class correlations on CLIP's segmentation capability remains unclear. On the one hand, inter-class correlations may reduce the discriminability of image patches by aggregating information from other categories. On the other, such correlations may facilitate the aggregation of essential semantic information in inter-class image patches, enabling better alignment with corresponding category embeddings.

To investigate the impact of inter-class correlations, we restrict the scope of patch correlations to intra-class. It significantly improves CLIP's segmentation performance, as shown in \cref{patch1}, demonstrating that intra-class correlations benefit CLIP's segmentation. We then gradually introduce inter-class correlations. Performance declines as the proportion of inter-class correlations increases. Even interactions with the most similar inter-class patches consistently lead to performance degradation in segmentation tasks (we use DINO \cite{dino} to measure the similarity, which is more semantically coherent than CLIP \cite{register, lost}). Similar phenomena on other datasets are presented in the supplementary. Additionally, intra-object correlations are sufficient to enhance CLIP's segmentation capability. This further demonstrates that inter-class correlations are the main reason for the limited segmentation performance of CLIP.

Existing representative methods \cite{sclip, proxyclip} modify self-attention to enhance CLIP's segmentation performance, which could be attributed to the reduction of inter-class correlations. As indicated in \cref{patch2}, the performance of these methods decreases as the proportion of inter-class correlations increases. SCLIP \cite{sclip} employs self-self attention, which reduces inter-class correlations because it makes patches focus on themselves. However, the incoherent CLIP's features still lead to many inter-class correlations. ProxyCLIP \cite{proxyclip} leverages the features of more coherent vision foundation models and applies thresholding to remove low-similarity inter-class correlations. Its performance is still limited because the similarity thresholding cannot reduce high-similarity inter-class correlations.

Therefore, it is necessary to explicitly regulate the scope of patch interactions to more effectively mitigate the adverse effects of inter-class correlations. To this end, we propose scope reconstruction to confine the scope of patch interactions within regions. Specifically, we use SAM \cite{sam, sam2} to get the region masks of the image and restrict the scope of patch interactions to these regions. As shown in \cref{patch2}, our scope reconstruction reduces the proportion of inter-class correlations and significantly improves the segmentation performance (the effectiveness on other datasets is demonstrated in \cref{sr2}). Because SAM-generated masks may have multiple classes, we employ value reconstruction to obtain more coherent similarity values, thereby further reducing the weight of inter-class correlations.

Apart from patch correlations, we propose feature refinement and map correction from the perspectives of feature representations and segmentation maps. Feature refinement introduces two additional branches to strengthen the patch features’ spatial details and semantic representation. Specifically, the spatial branch incorporates the lower layers' features into the final patch features to improve spatial details, and the semantic branch utilizes mask class tokens to aggregate global information within region masks to enhance semantic representation. Map correction leverages region masks to update segmentation results, improving spatial consistency. With the careful designs above, CorrCLIP significantly outperforms existing SOTA approaches on eight benchmarks, boosting the averaged mIoU from 48.6\% to 53.6\%.

Our contributions include: (1) We reveal that inter-class correlations are the primary reason for impairing CLIP's segmentation performance. Accordingly, we propose scope and value reconstruction to reduce inter-class correlations. (2) We introduce feature refinement to enhance the spatial details and semantic representation of patch features, along with map correction to improve the spatial consistency of segmentation results. (3) We demonstrate the superior performance of CorrCLIP across eight benchmarks.

\section{Related Work}

\textbf{Open-Vocabulary Semantic Segmentation.} Unlike traditional semantic segmentation \cite{fcn, mask2former, tang2025rethinking}, OVSS segments an image based on categories described by texts. Recent advancements in OVSS are primarily due to the development of large-scale vision-language models (VLMs) \cite{clip, openclip, metaclip}. OVSS methods can be divided into two categories: training-based and training-free. Training-based methods rely on mask annotations \cite{sed,cat-seg,yu2023convolutions,san,openvg,scan,use}, images \cite{clip-dinoier,reco}, or texts \cite{tcl, code, CoCu, PGSeg, segclip}. While training-based methods are typically more effective on specific datasets, they carry the potential risk of reducing the open-vocabulary capacity inherited from VLMs, as described in CaR \cite{car}. In contrast, training-free methods do not need any training and fully leverage the open-vocabulary capabilities of VLMs. Some methods \cite{maskclip,sclip,clip-srugery,lan2024clearclip, naclip} modify the attention mechanism in the final layer from query-key to self-self, enabling patches to focus more on themselves. Others \cite{resclip, sc-clip, cliper} observe that lower-layer attentions exhibit better semantic coherence and thus leverage them to refine the attention maps in the final layer. However, the performance of these approaches is constrained by the inherent limitations of CLIP's image-text alignment training objective as illustrated in \cref{radar}. Consequently, alternative methods \cite{proxyclip, cass, cliper, trident, freeda, lavg} utilize characteristics of other foundational models to enhance CLIP's segmentation capability. Yet, none of the existing methods effectively regulate the scope of patch interactions to mitigate the adverse effects of inter-class correlations. In this paper, we propose scope reconstruction to explicitly define the scope of patch interactions, significantly enhancing CLIP's segmentation performance.

\noindent\textbf{Vision-Language Pre-training.} Vision-language pre-training aims to enable models to learn cross-modal information through weakly supervised training on image-text pairs. This pre-training process allows models to understand the associations between images and corresponding texts. The performance of early research \cite{oscar, vilbert, unicoder} is constrained by the limited size of datasets. Recent studies \cite{clip, align, florence}, which leverage large-scale web data, have developed more robust representations. Among these, CLIP \cite{clip} stands out as the most popular vision-language model. It employs contrastive learning to align images with corresponding captions, achieving impressive generalization capabilities on unseen data. Subsequent research \cite{openclip, metaclip, eva-clip, sigclip, altogether, mode, dfnclip} further enhances CLIP by optimizing training data and processes.

\noindent\textbf{Vision Foundation Models.} Vision foundation models (VFMs) undergo pre-training on large-scale datasets to capture general feature representations of the visual world. VFMs learn rich underlying visual patterns, which can be fine-tuned or directly applied to various visual tasks. One type of VFMs is self-supervised models \cite{mae, dino, dinov2, ibot}, which aim to learn general-purpose visual features solely from images. Among these, the DINO \cite{dino} model shows the power to capture the semantic layout of images \cite{register, lost}. Another type of VFMs is the SAM \cite{sam} series, demonstrating impressive zero-shot, class-agnostic segmentation capabilities. Recent advancements have focused on improving the quality of the generated masks \cite{segment-hq, huang2024hrsam} and enhancing efficiency to enable broader applications in real-world and mobile scenarios \cite{efficientsam, fasterSAM, zhao2023fast, zhou2023edgesam, zhang2024efficientvit, sam2}. We capitalize on the strong generalization of SAM and DINO to effectively reduce inter-class correlations.

\section{Method}
\begin{figure*}
    \includegraphics[width=1\linewidth]{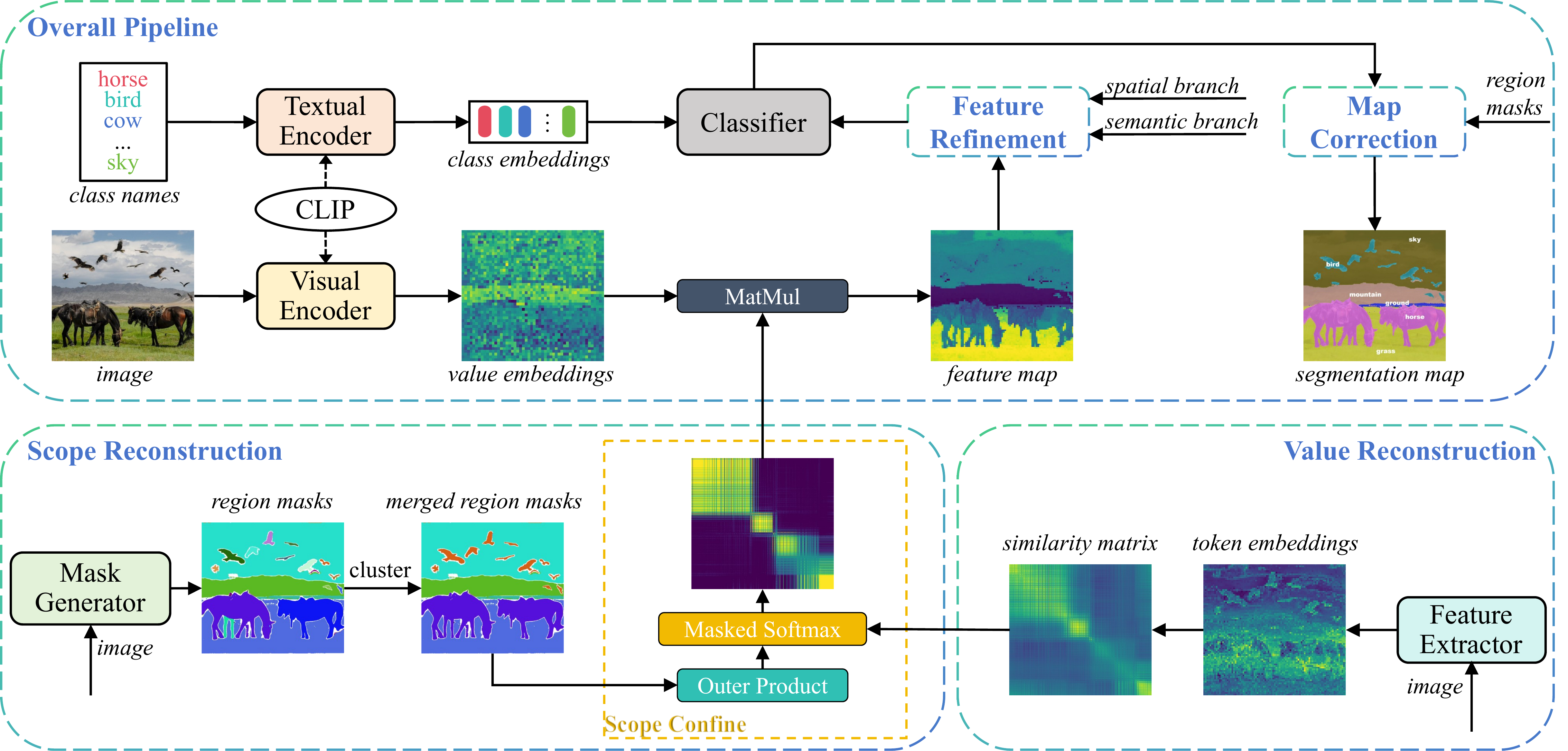}
    \caption{The overall framework of CorrCLIP. CLIP's visual encoder processes the image to obtain value embeddings, while CLIP's textual encoder encodes class names to produce class embeddings. A mask generator creates region masks, which are then clustered. A feature extractor derives semantically coherent token embeddings to compute the similarity matrix for value reconstruction. Applying the region masks to the similarity matrix achieves scope reconstruction. Through correlation reconstruction, value embeddings become more distinct. The feature map is refined via spatial and semantic branches, and the refined features are compared with class embeddings to generate the segmentation map, which is further updated using the region masks.}
  \label{framework}
\end{figure*}

In this section, we first introduce the overall process of adapting CLIP to OVSS in \cref{Preliminary}. Then, we introduce our scope reconstruction in \cref{Scope} and value reconstruction in \cref{Value}. We present feature refinement in \cref{Refinement} and map correction in \cref{Correction}. The overall framework is shown in \cref{framework}.

\subsection{Preliminary} \label{Preliminary}
In ViT, an image is first divided into patches, which are then transformed into token embeddings using a linear layer. These tokens are flattened to form the token sequence represented as $X_C \in \mathbb{R} ^ {N \times d}$, where $N$ is the sequence length, and $d$ is the dimension of tokens. Next, positional encoding is added to provide position information. The token sequence is then input into the CLIP's visual encoder. In each preceding layer before the final layer, the token sequence sequentially passes through a multi-head attention network and a feed-forward network, with residual connections applied after each sub-layer.

In the last layer, the token sequence $X_C$ is mapped to query, key, and value embeddings, denoted as $Q_C$, $K_C$, and $ V_C \in \mathbb{R} ^ {N \times d}$, respectively. For clarity of exposition, we only show single-head attention. Next, the similarity matrix $S \in \mathbb{R} ^ {N \times N}$ is computed by the dot product of query embeddings to improve the semantic coherence as illustrated in \cite{lan2024clearclip}:
\begin{equation}
    S = QQ^\mathrm{T} \label{similarity}
\end{equation}

$S$ represents the patch correlations, and its computation is the key to enhancing CLIP's segmentation capability. In the following subsections, we will introduce how to reconstruct patch correlations to improve segmentation performance. 
The attention map is derived from the similarity matrix by applying the softmax operation. Then, the attention map is applied to $V_C$ to aggregate information from all patches in the input sequence. To reduce the detrimental impact of noise on the segmentation results, as illustrated in ClearCLIP \cite{lan2024clearclip}, we remove the residual connections and the feed-forward neural network in the last layer. Subsequently, the sequence of tokens is projected to the final image patch features $F_{img} \in \mathbb{R} ^ {N \times d}$:
\begin{align}
    Attn &= \mathrm{Softmax}(\frac{S}{\sqrt{d}}) \label{attn}\\ 
    F_{img} &= \mathrm{Proj}(Attn V_C)
    \label{attn_value}
\end{align}

Meanwhile, $K$ class names are combined with standard ImageNet prompts \cite{clip} to form category descriptions, which are then encoded by CLIP's textual encoder to generate class embeddings $F_{text} \in \mathbb{R} ^ {K \times d}$. Then, image patch features are projected to align with the feature space of class embeddings. Finally, the class embeddings serve as the classifier's parameters to generate the segmentation map:
\begin{equation}
    pred = \mathop{\arg\max}\limits_{K}(\mathrm{Proj}(F_{img}) F_{text}^\mathrm{T})
\end{equation}

\subsection{Scope Reconstruction} \label{Scope}
Although the self-self attention can enhance the segmentation performance, it is still limited to CLIP's incoherent patch embeddings, which leads to many inter-class correlations. To address this, we reconstruct the scope of patch interactions using SAM.

SAM uniformly samples points in the image and generates masks for each point. Then thresholding is applied to these masks to discard those with lower confidence and stability. Subsequently, these masks are downsampled to match the size of the final feature map in CLIP. Consequently, we obtain $Z$ non-overlapping region masks $M = \{m_1, m_2, ..., m_Z\} \in \mathbb{R} ^ {Z \times N}$, where $m_i \in \mathbb{R} ^ N$ represents the $i$th region mask. The union of all unsegmented regions is denoted by $m_0$.

We merge similar region masks to achieve a closer alignment with the true class masks. Specifically, we acquire the features of all segmented regions $F_{region} = \{f_1, f_2, ..., f_Z\} \in \mathbb{R} ^ {Z \times d}$ by performing mask average pooling on the token embeddings $F_S$. Subsequently, we use a clustering algorithm to merge similar regions based on these region features:
\begin{align}
    f_i &= \mathrm{Mean}(m_i \odot F_S) \\
    \hat{M} &= \mathrm{Cluster}(M, F_{region})
\end{align}
where $\hat{M} = \{\hat{m}_1, \hat{m}_2, ..., \hat{m}_z\} \in \mathbb{R} ^ {z \times N}$ are merged region masks and $z$ is the number of merged region masks. $\odot$ represents the element-wise product. Note that merging masks may result in masks containing multiple classes. However, we believe these classes are highly similar, so the impact of inter-class correlations is minimal, meaning that the benefits of merging masks outweigh the potential drawbacks. We empirically validated this in \cref{mask merging}, observing improved segmentation performance across all datasets.

We then use these merged region masks to calculate the interaction matrix $E \in \mathbb{R} ^ {N \times N}$ where $E_{i,j} = 1$ means that the $i$th patch can attend to the $j$th patch. For patches in unsegmented regions, we set the average value of the similarity matrix as a threshold, allowing only patches with similarity values exceeding this threshold to interact:
\begin{equation}
    E = \sum_{i=1}^z\hat{m}_i \otimes \hat{m}_i + (m_0 \otimes m_0) \odot (S > \mathrm{Mean}(S))
    \label{semantic-matrix}
\end{equation}
where \cref{semantic-matrix} employs the broadcasting mechanism, and $\otimes$ denotes outer product. Finally, we restrict the scope of patch interactions by applying the masked softmax, where $E$ represents the mask. \cref{attn} is changed as follows:

\begin{equation}
Attn = \mathrm{Masked~Softmax}(\frac{S}{\sqrt{d}},~E) \label{masked attention}
\end{equation}

\subsection{Value Reconstruction} \label{Value} 
Some generated masks may contain multiple classes, causing inter-class correlations. To mitigate the adverse effects of inter-class correlations, we harness DINO's understanding of semantic layout to construct a more coherent similarity matrix. This could reduce the value of inter-class correlations. Inspired by previous work \cite{sclip,clip-srugery,cliptrase}, which combines query and key embeddings in CLIP, we utilize this combination to fully leverage DINO's capabilities.

Like CLIP, the token sequence in DINO's final layer is mapped to $Q_D$, $K_D$, and $V_D \in \mathbb{R} ^ {N \times d}$, respectively. In practice, the token sequence lengths of CLIP and DINO may differ due to the variation in patch size, which can be addressed through interpolation. The similarity matrix in \cref{similarity} and attention map in \cref{masked attention} are changed as follows:
\begin{align}
    S = \frac{F_S F_S^\mathrm{T}}{\Vert F_S \Vert^2} =& \frac{(Q_D + K_D)(Q_D + K_D)^\mathrm{T}}{\Vert Q_D + K_D \Vert^2} \\
    Attn = \mathrm{Mas}&\mathrm{ked~Softmax}(\frac{S}{\tau},~E)
\end{align}
where $\tau < 1$ is a temperature coefficient used to sharpen the attention distribution, amplifying the scores for highly similar patches.

\subsection{Feature Refinement} \label{Refinement}
After reconstructing patch correlations, attention is applied to the value embeddings to generate the final patch features. We define this core process as our main branch and propose two additional branches to refine the final patch features. 

We present mask class tokens to enhance CLIP's class token benefits for OVSS. Specifically, we repeat $z$ distinct class tokens at the start of ViT, each corresponding to one mask. In the attention layers, each mask class token exclusively attends to the image tokens within its corresponding mask to capture mask-specific semantics. At the end of ViT, each image token within a mask is summed with its corresponding mask class token. The final addition enriches the semantic information of the image tokens within each mask, facilitating better alignment with the corresponding text vectors. Additionally, building on recent studies \cite{sc-clip, resclip, cliper} that CLIP's lower-layer patch features retain richer spatial information, we incorporate these features into the final patch features to enhance spatial details. Finally, image features in \cref{attn_value} are changed as follows:
\begin{equation}
F_{img} = \mathrm{Proj}(AttnV_C) + \alpha * \mathrm{Proj}(AttnV_C') + \beta * MCT
\end{equation}
where $V_C'$ is the lower layers’ features and is mapped by the final layer’s value projection inspired by \cite{sc-clip}. Further discussion on the spatial branch is presented in the supplementary. $MCT$ represents the mask class tokens. $\alpha$ and $\beta$ are the coefficients that balance these two additional branches. Empirically, $\alpha$ is set to 1, and $\beta$ is set to 0.5.

\subsection{Map Correction} \label{Correction}
Spatial consistency is essential in segmentation tasks, referring to maintaining continuity in space. Simply put, it means ensuring that the predictions of adjacent pixels are logically consistent, avoiding abrupt changes such as predicting other objects within a coherent object. Fully supervised methods can learn consistency with mask annotations, but CLIP lacks this capability due to its weak supervision. Although our correlation reconstruction alleviates this issue, it still falls short of the performance achieved with full supervision. To bridge this gap, we reuse the region masks generated above to update the segmentation map. Specifically, we change the categories of all patches within a region to the category that is most common within this region:
\begin{equation}
    pred[{m}_i] = \mathrm{Mode}(pred[{m}_i]),~~~~i > 0
\end{equation}

\section{Experiments}
\begin{table*}
  \centering
  \resizebox{\textwidth}{!}{
  \begin{tabular}{lc|llllllll|l}
    \toprule
    Method & Size & VOC21 & VOC20 & PC60 & PC59 & Object & Stuff & ADE & City & Avg\\
    
    \midrule
    \rowcolor{ForestGreen!8}
    \multicolumn{11}{c}{\textbf{\emph{Training-based}}}  \\
    TCL\cite{tcl} & \multirow{5}{*}{ViT-B/16} & 55.0 & 83.2 & 30.4 & 33.9 & 31.6 & 22.4 & 17.1 & 24.0 & 37.2\\
    CLIP-DINOiser\cite{clip-dinoier} &  & 62.1 & 80.9 & 32.4 & 35.9 & 34.8 & 24.6 & 20.0 & 31.7 & 40.3\\
    CoDe\cite{code}&  & 57.7 & - & 30.5 & - & 32.3 &  23.9 & 17.7 & 28.9 & -\\

    CAT-Seg\cite{cat-seg} &  & 77.3 & 94.6 & - & 57.5 & - & - & 31.8 & - & -\\
    ESC-Net\cite{esc-net} & & 80.1 & 97.3 & - & 59.0 & - & - & 35.6 & - & -\\
    
    \midrule
    \rowcolor{ForestGreen!8}
    \multicolumn{11}{c}{\textbf{\emph{Training-free}}}   \\
    CLIP\cite{clip}& \multirow{14}{*}{ViT-B/16} & 11.5 & 41.9 & 4.4 & 9.2 & 1.6 & 4.4 & 2.9 & 5.0 & 10.1\\
    MaskCLIP\cite{maskclip}& & 38.8 & 74.9 & 23.6 & 26.4 & 20.6 & 16.4 & 9.8 & 12.6 & 27.9\\
    ClearCLIP\cite{lan2024clearclip}& & 51.8 & 80.9 & 32.6 & 35.9 & 33.0 & 23.9 & 16.7 & 30.0 & 38.1\\
    SCLIP\cite{sclip}& & 59.1 & 80.4 & 30.4 & 34.2 & 30.5 & 22.4 & 16.1 & 32.2 & 38.2\\
    ProxyCLIP\cite{proxyclip}&  & 61.3 & 80.3 & 35.3 & 39.1 & 37.5 & 26.5 & 20.2 & 38.1 & 42.3\\
    LaVG\cite{lavg}&&  62.1 & 82.5 &  31.6 & 34.7 &  34.2 & 23.2 & 15.8 & 26.2 & 38.8 \\
    CLIPtrase\cite{cliptrase}& & 53.0 & 81.2 & 30.8 & 34.9 & \textbf{44.8} & 24.1 & 17.0 & -    & -\\
    NACLIP\cite{naclip} & &64.1 & 83.0 &  35.0 & 38.4 & 36.2 &  25.7 & 19.1 & 38.3 & 42.5\\
    Trident\cite{trident} & &\underline{67.1} & 84.5 & \underline{38.6} & \underline{42.2} & 41.1 & \underline{28.3} & \underline{21.9} & \underline{42.9} & \underline{45.8}\\
    ResCLIP\cite{resclip}&& 61.1 & 86.0 & 33.5 & 36.8 & 35.0 & 24.7 & 18.0 & 35.9 & 41.4 \\
    SC-CLIP\cite{sc-clip}&& 64.6 & 84.3 & 36.8 & 40.1 & 37.7 & 26.6 & 20.1 & 41.0 & 43.9 \\
    CLIPer\cite{cliper}&& 65.9 & 85.2 & 37.6 & 41.7 & 39.0 & 27.5 & 21.4 & - & - \\
    CASS\cite{cass}&& 65.8 & \underline{87.8} & 36.7 & 40.2 & 37.8 & 26.7 & 20.4 & 39.4 & 44.4 \\
    \rowcolor{gray!8}
    CorrCLIP (Ours)& & \textbf{74.8} \textcolor{blue}{(+7.7)} & \textbf{88.8} \textcolor{blue}{(+1.0)} & \textbf{44.2} \textcolor{blue}{(+5.6)} & \textbf{48.8} \textcolor{blue}{(+6.6)}& \underline{43.7} \textcolor{blue}{(-1.1)}& \textbf{31.6} \textcolor{blue}{(+3.3)}& \textbf{26.9} \textcolor{blue}{(+5.0)}& \textbf{49.4} \textcolor{blue}{(+6.5)}& \textbf{51.0} \textcolor{blue}{(+5.2)}\\
    
    \midrule
    FreeDA\cite{freeda}& \multirow{7}{*}{ViT-L/14} & 55.4 & 87.9 & \underline{38.3} & 43.5 & 37.4 & \underline{28.8} & 23.2 & 36.7 & 43.9\\
    CaR\cite{car}&  & 67.6 & \underline{91.4} & 30.5 & 39.5 & 36.6 & - & 17.7 & - & -\\
    ProxyCLIP\cite{proxyclip}&  & 60.6 & 83.2 & 34.5 & 37.7 & 39.2 & 25.6 & 22.6 & 40.1 & 43.0\\
    ResCLIP\cite{resclip}&& 54.1 & 85.5 & 30.9 & 34.5 & 32.5 & 23.4 & 18.2 & 33.7 & 39.1 \\
    SC-CLIP\cite{sc-clip}&& 65.0 & 88.3 & 36.9 & 40.6 & 40.5 & 26.9 & 21.7 & \underline{41.3} & \underline{45.2} \\
    CLIPer\cite{cliper}&& \underline{69.8} & 90.0 & 38.0 & \underline{43.6} & \underline{43.3} & 28.7 & \underline{24.4} & - & - \\
    \rowcolor{gray!8}
    CorrCLIP (Ours)&  & \textbf{76.7} \textcolor{blue}{(+6.9)}& \textbf{91.5} \textcolor{blue}{(+0.1)}& \textbf{44.9} \textcolor{blue}{(+6.6)}& \textbf{50.8} \textcolor{blue}{(+7.2)}& \textbf{49.4} \textcolor{blue}{(+6.1)}& \textbf{34.0} \textcolor{blue}{(+5.2)}& \textbf{30.7} \textcolor{blue}{(+6.3)}& \textbf{51.1} \textcolor{blue}{(+9.8)}& \textbf{53.6} \textcolor{blue}{(+8.4)}\\
    
    \midrule
    ProxyCLIP\cite{proxyclip}& \multirow{3}{*}{ViT-H/14} & 65.0 & 83.3 & 35.4 & 39.6 & 38.6 & 26.8 & 24.2 & 42.0 & 44.4\\
    Trident\cite{trident} & &\underline{70.8} & \underline{88.7} & \underline{40.1} & \underline{44.3} & \underline{42.2} & \underline{28.6} & \underline{26.7} & \underline{47.6} & \underline{48.6}\\
    \rowcolor{gray!8}
    CorrCLIP (Ours)& & \textbf{76.4} \textcolor{blue}{(+5.6)}& \textbf{91.8} \textcolor{blue}{(+3.1)}& \textbf{42.5} \textcolor{blue}{(+2.4)}& \textbf{47.9} \textcolor{blue}{(+3.6)}& \textbf{48.4} \textcolor{blue}{(+6.2)}& \textbf{32.7} \textcolor{blue}{(+4.1)}& \textbf{28.8} \textcolor{blue}{(+2.1)}&\textbf{49.9} \textcolor{blue}{(+2.3)}& \textbf{52.3} \textcolor{blue}{(+3.7)}\\
    \bottomrule
  \end{tabular}
  }
  \caption{Comparison with state-of-the-art OVSS methods on eight benchmarks in three different sizes of CLIP. Bold fonts indicate the optimal methods and underlined fonts indicate the suboptimal methods. ``Avg'' represents the averaged mIoU across eight benchmarks.} 
  \label{sotas}
\end{table*}

\subsection{Dataset and Evaluation Metric} 
Following prior work, we evaluate our method on the validation sets of five datasets. \textbf{Pascal VOC} \cite{pascal2012} comprises 1,449 images and serves two benchmarks: VOC21 (21 classes with background) and VOC20 (20 classes without). \textbf{Pascal Context} \cite{context} contains 5,104 images and similarly supports PC60 (60 classes) and PC59 (59 classes). \textbf{COCO Stuff} \cite{coco} (Stuff) includes 5,000 images divided into 171 classes, which encompass both stuff and object categories. COCO Object (Object) is a derivative of COCO Stuff, which merges all stuff classes into the background class and has 81 classes. \textbf{ADE20k} \cite{ade} (ADE) has 2,000 images and 150 classes without the background class. \textbf{Cityscapes} \cite{cityscapes} (City) has 500 images and 19 classes without the background class. These datasets form eight benchmarks (5 without and 3 with background classes). We compare results using mean Intersection over Union (mIoU).
\subsection{Implementation Details}
Our model has three different sizes based on CLIP: ViT-B/16 \cite{metaclip}, ViT-L/14 \cite{metaclip}, and ViT-H/14 \cite{openclip}. All other configurations for the three variants are identical. The backbone of DINO is ViT-B/8 \cite{dino}. We use SAM2 \cite{sam2} with MAE \cite{mae} pre-trained Hiera-L \cite{ryali2023hiera, bolya2023window}. Results for other mask generators are presented in the supplementary. We collect 32$\times$32 prompt points in a grid manner to generate region masks of an image.  ``pred\_iou\_thresh'' and ``stability\_score\_thresh'' in mask thresholding are set to 0.7 for all datasets. The clustering algorithm used in mask merging is DBSCAN \cite{dbscan}, where the neighborhood radius is set to 0.2, and the minimum number of samples is set to 1 for all datasets. The temperature coefficient $\tau$ is set to 0.25.

We resize the images to meet the varying specifications of different datasets: a shorter side of 336 pixels for Pascal VOC, Pascal Context, and COCO, and 448 pixels for Cityscapes and ADE20K. We perform slide inference with a 336$\times$336 window and 112$\times$112 stride.

\begin{figure*}[t]
  \centering
   \includegraphics[width=1\linewidth]{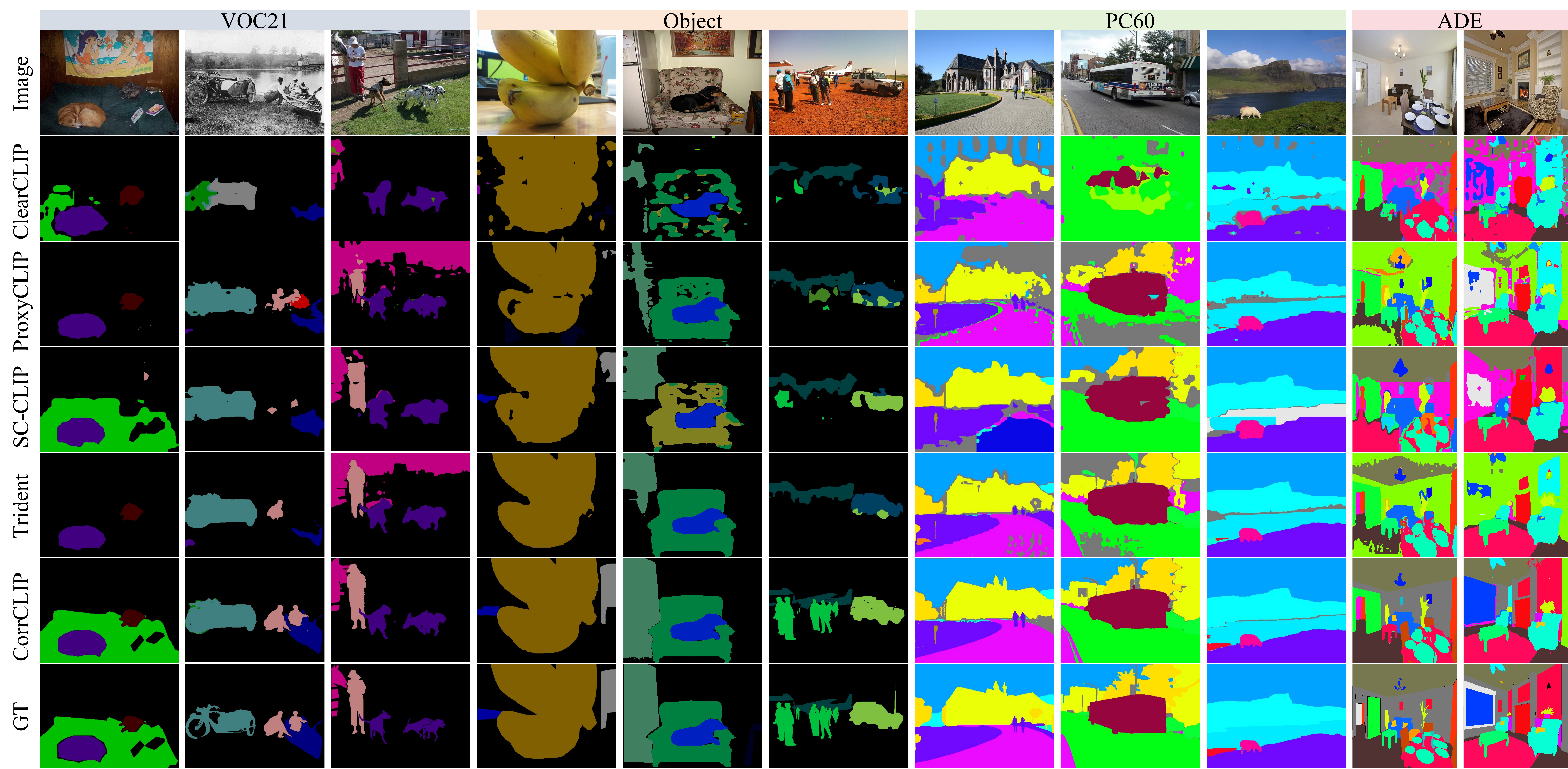}
   \caption{Qualitative comparison between our method CorrCLIP and the other four methods. ``GT'' denotes ground truth.}
   \label{qualitative comparison}
\end{figure*}
\subsection{Comparison with State-of-the-Art Methods}

We compare CorrCLIP with state-of-the-art training-free OVSS methods across eight benchmarks. In these methods, MaskCLIP \cite{maskclip}, CaR \cite{car}, SCLIP \cite{sclip}, ClearCLIP \cite{lan2024clearclip}, CLIPtrase \cite{cliptrase}, NACLIP \cite{naclip}, SC-CLIP\cite{sc-clip}, and ResCLIP \cite{resclip} only use CLIP. The incoherent patch features of CLIP constrain their performance. LaVG \cite{lavg}, ProxyCLIP \cite{proxyclip}, Trident \cite{trident}, CLIPer \cite{cliper}, CASS \cite{cass}, and FreeDA \cite{freeda} use foundation models to alleviate CLIP's limitation. However, they do not effectively confine the scope of correlations. 

The quantitative results are summarized in \cref{sotas}. All three variants of our CorrCLIP demonstrate superior performance across eight benchmarks, significantly outperforming other methods. In particular, CorrCLIP achieves increases of 5.2\%, 8.4\%, and 3.7\% in averaged mIoU across three variants. We also compare our approach with several state-of-the-art training-based methods \cite{tcl, clip-dinoier, code, cat-seg, esc-net}. CorrCLIP surpasses weakly-supervised methods and narrows the gap to fully-supervised ones. It further demonstrates superior generalization by outperforming fully-supervised models on out-of-distribution datasets, as detailed in the supplementary.

In \cref{qualitative comparison}, we provide qualitative comparisons of our method, CorrCLIP, with ClearCLIP \cite{lan2024clearclip}, ProxyCLIP \cite{proxyclip}, SC-CLIP \cite{sc-clip}, and Trident \cite{trident}. Due to the explicitly restricted interaction scope, our method correctly identified objects misclassified by other approaches. Our method also demonstrates strong object continuity and significantly reduces noise compared to other methods. Additional qualitative comparisons are provided in the supplementary.

\begin{table}[t]
  \centering
    \small
    \setlength{\belowcaptionskip}{-5pt}
  \renewcommand{\arraystretch}{0.80}
  \begin{tabular}{@{}p{0.3cm}<{\centering}p{0.3cm}<{\centering}p{0.3cm}<{\centering}p{0.3cm}<{\centering}|ccccc@{}}
    \toprule
    SR & VR & MC & FR & VOC21 & PC60 & Object & ADE &City\\
    \midrule
     &  & & & 53.6 & 31.9 & 32.0 & 19.3 & 30.5\\
    \checkmark& & & & 68.1 & 39.6 & 41.5 & 24.0 & 42.0  \\
     \checkmark& \checkmark & & & 68.5 & 40.3 & 42.0 & 24.0 & 43.2\\
    \checkmark& \checkmark & \checkmark & & 72.5 & 42.0 & 43.7 & 25.3 & 48.3\\
     \checkmark& \checkmark & \checkmark & \checkmark & \textbf{74.8} & \textbf{44.2} & \textbf{43.7} & \textbf{26.9} & \textbf{49.4}\\
    \bottomrule
  \end{tabular}
  \caption{Impact of integrating different components.}
  \label{component}
\end{table}

\subsection{Ablation Study}
We perform a series of ablation studies to investigate the effects of each component in our method.

\noindent \textbf{Impact of Integrating Different Components.} We investigate the impact of four components of CorrCLIP outlined in \cref{component}. These components are scope reconstruction (SR), value reconstruction (VR), map correction (MC), and feature refinement (FR). We use ClearCLIP \cite{lan2024clearclip} as the baseline, which substitutes query-key with query-query and removes the feed-forward network and residual connections. SR significantly enhances CLIP's segmentation performance, yielding improvements of 14.5\%, 7.7\%, 9.5\%, 4.7\%, and 11.5\% mIoU on VOC, PC, Object, ADE, and City, respectively. VR has a further improvement. Incorporating MC and FR also leads to performance gains. These results demonstrate the effectiveness of each component. The computational costs of each component and the corresponding analyses are presented in the supplementary.

\begin{table}[t]
  \centering
  \small
  \setlength{\belowcaptionskip}{-5pt}
  \renewcommand{\arraystretch}{0.80}
  \begin{tabular}{@{}p{1.8cm}<{} | p{0.73cm}<{} p{0.73cm}<{} p{0.73cm}<{} p{0.73cm}<{} p{0.73cm}<{}@{}}
    \toprule
    Method & VOC21 & PC60 & Object& ADE  & City\\
        \midrule
    SCLIP\cite{sclip}& 59.6 & 31.7 & 33.5 & 16.5  & 32.3\\
    \rowcolor{gray!12}
    +SR & 
    66.5\textsubscript{\textcolor{blue}{+6.9}} & 
    35.9\textsubscript{\textcolor{blue}{+4.2}} & 
    37.0\textsubscript{\textcolor{blue}{+3.5}} & 
    19.8\textsubscript{\textcolor{blue}{+3.3}} & 
    39.0\textsubscript{\textcolor{blue}{+6.7}} \\
    \midrule
    ProxyCLIP\cite{proxyclip}& 61.3 & 35.3 & 37.5 & 20.2 & 38.1\\
    \rowcolor{gray!12}
    +SR & 
      67.0\textsubscript{\textcolor{blue}{+5.7}} & 
      38.2\textsubscript{\textcolor{blue}{+3.1}} & 
      41.7\textsubscript{\textcolor{blue}{+4.2}} & 
      21.6\textsubscript{\textcolor{blue}{+1.4}} &
      40.8\textsubscript{\textcolor{blue}{+2.7}} \\
    \midrule
    SC-CLIP\cite{sc-clip}& 64.6 & 36.8 & 37.7& 20.1  & 41.0\\
    \rowcolor{gray!12}
    +SR & 
    66.0\textsubscript{\textcolor{blue}{+1.4}} & 
    37.9\textsubscript{\textcolor{blue}{+1.1}} & 
    38.8\textsubscript{\textcolor{blue}{+1.1}} & 
    20.9\textsubscript{\textcolor{blue}{+0.8}} & 
    42.4\textsubscript{\textcolor{blue}{+1.4}} \\
    \bottomrule
  \end{tabular}
  \caption{Effectiveness of Scope Reconstruction on other methods.}
  \label{sr1}
\end{table}

\noindent \textbf{Effectiveness of Scope Reconstruction.} Our proposed scope reconstruction can be seamlessly integrated into other methods \cite{sclip, proxyclip, sc-clip}. As shown in \cref{sr1}, our SR can further enhance the performance of these methods, as they do not explicitly confine the scope of correlations. Additionally, a key limitation of the SR process is the required downsampling of generated masks to match the low-resolution patch feature map, which can introduce quantization errors. We hypothesize that mitigating this error by increasing the mask resolution would improve performance. To validate this, we upsample the patch feature map via interpolation to generate larger final masks. As shown in \cref{sr2}, the results confirm our hypothesis: SR performance steadily improves with mask size, underscoring the benefit of reducing downsampling errors. However, since upsampling is highly time- and memory-intensive, we set the final mask size to 42$\times$42.

\noindent \textbf{Ablation Study on Value Reconstruction.} We explore the effect of different features used to compute the similarity values after SR. As shown in \cref{vr}, our proposed combination of DINO's query and key embeddings yields the best average performance. Note that features from smaller DINO (ViT-S) also show comparable performance. Interestingly, we found that even without computing similarity (i.e., uniform similarity across all patches), our method still achieves strong performance. Meanwhile, CLIP's original query-key computation mechanism yields inferior results due to its generation of highly incoherent similarity values. However, it still outperforms existing SOTA methods, further validating the effectiveness of our proposed SR.

\begin{table}[t]
  \centering
  \small
  \renewcommand{\arraystretch}{0.80}
  \begin{tabular}{@{}c | ccccc@{}}
    \toprule
    Mask Size & VOC21 & PC60 & Object& ADE & City \\
    \midrule
    No SR & 53.6 & 31.9 & 32.0 & 19.3 & 30.5\\
    21$\times$21 & 64.7 & 37.8 & 39.5 & 22.8 & 38.4 \\
        \rowcolor{gray!12}
    42$\times$42 & 68.1 & 39.6 & 41.5 & 24.0 & 42.0 \\
    63$\times$63 & 69.3 & 40.1 & 42.3 & 24.4 & 43.3 \\
    84$\times$84 & \textbf{70.2} & \textbf{40.7} & \textbf{43.0} & \textbf{24.8} & \textbf{44.7} \\
    \bottomrule
  \end{tabular}
  \caption{Impact of the final masks' size on scope reconstruction.}
  \label{sr2}
\end{table}

\begin{table}[t]
  \centering
    \small
  \renewcommand{\arraystretch}{0.80}
  \begin{tabular}{@{}lc|ccccc@{}}
    \toprule
    Sim & Model & VOC21 & PC60  & Object& ADE & City\\
    \midrule
    Unif  & -      & 73.4 & 43.4 & 43.8& 26.6 & 47.0\\
    Q-K   & CLIP-B & 69.5 & 41.6 & 42.4& 25.7 & 45.1\\
    Q-Q   & CLIP-B & 73.8 & 43.4 & \textbf{44.3}& 26.7 & 48.0\\
    X-X   & DINO-B & 73.4 & 43.8 & 43.1& \textbf{26.9} & 49.0\\
        \rowcolor{gray!12}
    QK-QK & DINO-S & \textbf{74.3} & 43.9 & 43.7& 26.8 & 48.3\\
        \rowcolor{gray!12}
    QK-QK & DINO-B & 74.2 & \textbf{44.0} & 43.4& 26.8 & \textbf{49.1}\\
    \bottomrule
  \end{tabular}
  \caption{Effect of different features used to compute the similarity values after scope reconstruction. ``Unif'' denotes all patches have the same similarity. ``X'' denotes output features.}
  \label{vr}
\end{table}

\begin{table}[t]
  \centering
  \small
\setlength{\belowcaptionskip}{-10pt}
  \renewcommand{\arraystretch}{0.80}
  \begin{tabular}{@{}c|ccccc@{}}
    \toprule
    Cluster & VOC21 & PC60 & Object & ADE  & City\\
        \midrule
    $\times$& 74.2 & 44.0 & 43.4 & 26.8& 49.1 \\
    \checkmark& \textbf{74.8} & \textbf{44.2} & \textbf{43.7} & \textbf{26.9}& \textbf{49.4} \\
    \bottomrule
  \end{tabular}
  \caption{Effectiveness of mask merging.}
  \label{mask merging}
\end{table}

\noindent \textbf{Effectiveness of Mask Merging.} We explore the effectiveness of mask merging on multiple benchmarks. As shown in \cref{mask merging}, mask merging boosts segmentation performance across all datasets, as enhanced intra-class correlations outweigh the adverse effects of inter-class correlations. This is reasonable because mask merging generally introduces high-similarity inter-class correlations, which have less adverse effects on CLIP's segmentation performance.

\noindent \textbf{Ablation Study on CLIP.} We investigate the effect of different types and sizes of CLIP models on CorrCLIP. We focus on four specific types of CLIP: CLIP \cite{clip}, OpenCLIP \cite{openclip}, MetaCLIP \cite{metaclip}, and DFNCLIP \cite{dfnclip}. These models share the same architecture and only differ from training data and processes. As shown in \cref{clip type}, the segmentation performance of the original CLIP and ClearCLIP (our adopted baseline) does not improve with the enhancement of CLIP's zero-shot classification capability. In contrast, our method's segmentation performance is positively correlated with CLIP's zero-shot capability, demonstrating that our approach fully exploits the potential of CLIP.

\noindent \textbf{Ablation Study on Additional Branches.} We investigate the roles of our proposed two additional branches, as shown in \cref{branch}. The semantic and spatial branches improve segmentation performance on most datasets. However, their individual application leads to performance degradation on the City and Object datasets, as modifying the final patch features may disrupt the alignment between patches and category embeddings. Combining the two branches synergistically enhances their effects, guaranteeing consistent performance improvements across all evaluated datasets.

\begin{table}[t]
  \centering
  \small
  \renewcommand{\arraystretch}{0.80}
  \begin{tabular}{@{}lcc|ccc@{}}
    \toprule
    Type & Size & Acc & Plain & ClearCLIP & CorrCLIP\\
    \midrule 
    CLIP &\multirow{4}{*}{\makecell{ViT-B}}&68.3& 12.7& 37.8& 47.0\\
    OpenCLIP &&70.2& \textbf{13.9}& 38.6& 48.1\\
    MetaCLIP &&72.1&11.0&37.7 & 49.1\\
    DFNCLIP &&\textbf{76.2}&12.6&\textbf{39.4} & \textbf{49.5}\\
    \midrule
    CLIP &\multirow{4}{*}{\makecell{ViT-L}}&75.5& 5.3&\textbf{35.2} & 49.0 \\
    OpenCLIP &&75.3& \textbf{11.3}&32.1 & 48.7\\
    MetaCLIP &&79.2&4.6& 35.1& \textbf{51.6}\\
    DFNCLIP &&\textbf{81.4}&2.6& 34.6& 50.7\\
    \midrule
    OpenCLIP &\multirow{3}{*}{\makecell{ViT-H}}&78.0&\textbf{7.5}& \textbf{35.5}& 50.2\\
    MetaCLIP &&80.5&3.2& 24.0& 50.2\\
    DFNCLIP &&\textbf{83.4}&1.3& 34.4& \textbf{50.6}\\
    \bottomrule
  \end{tabular}
  \caption{Impact of varied CLIP on the performance of different OVSS methods. OVSS performance is evaluated on five benchmarks without the background class. ``Acc'' is the zero-shot accuracy on ImageNet \cite{imagenet}.}
  \label{clip type}
\end{table}

\begin{table}[t]
  \centering
    \small
    \setlength{\belowcaptionskip}{-5pt}
  \renewcommand{\arraystretch}{0.80}
  \begin{tabular}{@{}p{0.85cm}<{\centering}p{0.75cm}<{\centering}|ccccc@{}}
    \toprule
    Semantic & Spatial & VOC21 & PC60 & Object & ADE &City\\
    \midrule
    && 72.5 & 42.0 & 43.7 & 25.3 & 48.3\\
    \checkmark&& 74.1 & 43.2 & \textbf{44.0} & 26.5 & 47.6\\
    &\checkmark& 72.9 & 43.3 & 42.6 & 26.4 & \textbf{49.9}\\
    \checkmark&\checkmark& \textbf{74.8} & \textbf{44.2} & 43.7 & \textbf{26.9} & 49.4\\
    \bottomrule
  \end{tabular}
  \caption{Effectiveness of the spatial branch and semantic branch.}
  \label{branch}
\end{table}

\section{Conclusion}
In this paper, we reveal that inter-class correlations significantly impair CLIP's segmentation performance. Accordingly, we propose scope reconstruction to confine the scope of patch interactions, reducing inter-class correlations. To alleviate the problem that scope reconstruction cannot completely eliminate inter-class correlations, we present value reconstruction to compute more coherent similarity values, reducing the weight of inter-class correlations. Moreover, we introduce feature refinement to enhance the spatial granularity and semantic richness of patch features and map correction to improve the spatial consistency of segmentation maps. Armed with the above designs, our method significantly enhances CLIP's segmentation capability and demonstrates superior performance across eight benchmarks.

\clearpage
{
    \small
    \bibliographystyle{ieeenat_fullname}
    \bibliography{main}
}
\clearpage
\maketitlesupplementary

\begin{figure*}[htbp]
  \centering
   \includegraphics[width=1\linewidth]{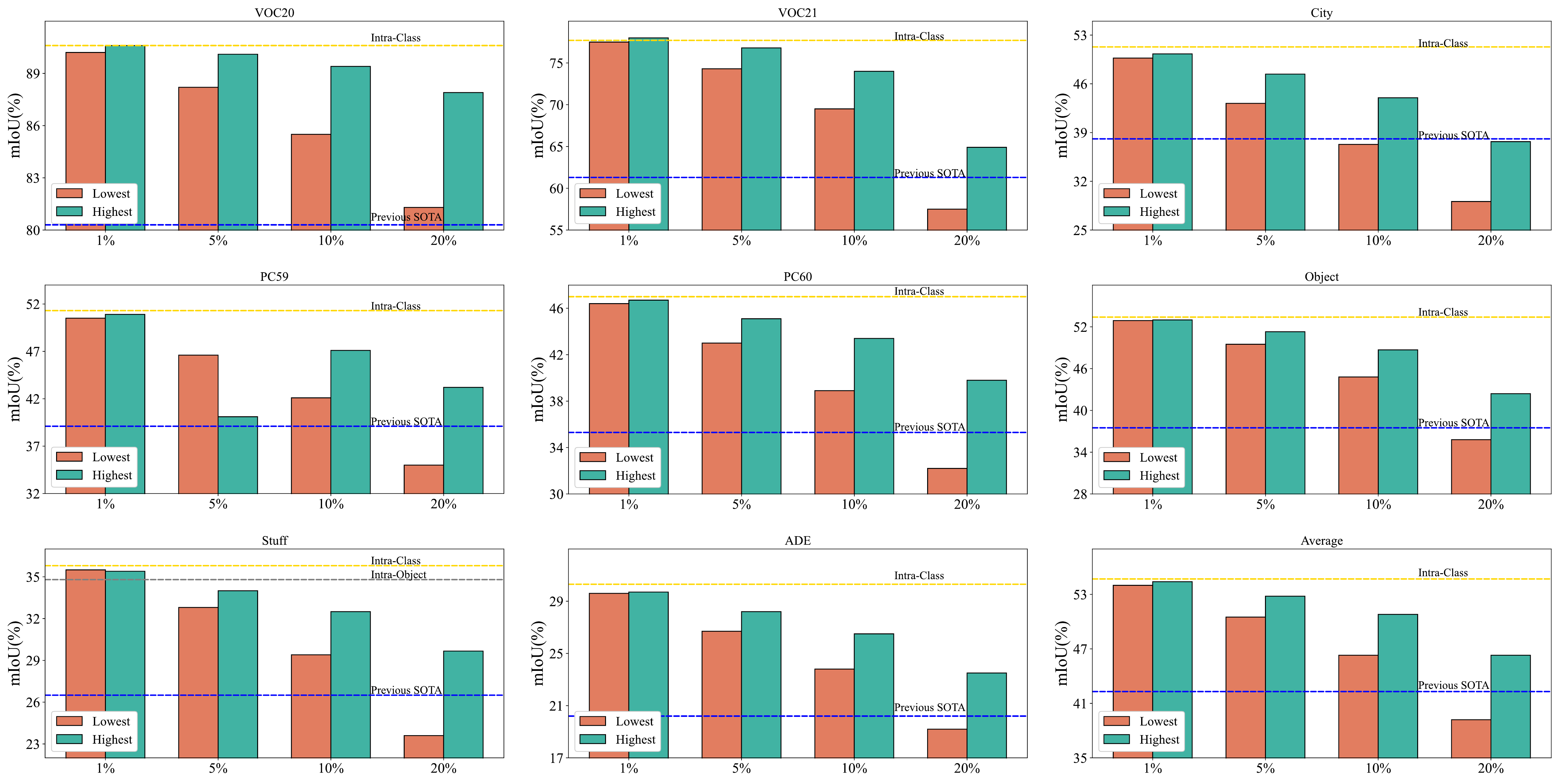}
   \caption{Impact of intra-class and inter-class correlations on CLIP's segmentation capability on the other seven benchmarks and the average of all benchmarks.}
   \label{patch_supp}
\end{figure*}

\textbf{Intra-Class and Inter-Class Correlations.} We explore the impact of inter- and intra-class correlations on CLIP's segmentation capabilities on other datasets, as shown in \cref{patch_supp}. Consistent with the findings in the main text, intra-class correlations can significantly enhance CLIP's segmentation performance, while inter-class correlations impair its segmentation capability.

\textbf{Performance on Out-of-Distribution Datasets.} Fully-supervised methods \cite{cat-seg, openvg} are typically trained on the COCO dataset, which shares high similarity with the five datasets used in the main text. This setup, however, does not adequately showcase the full potential of OVSS. Therefore, we conduct additional evaluations on a selection of datasets from the MESS benchmark \cite{mess} that deviate from the COCO distribution. As shown in \cref{mess}, CorrCLIP outperforms the fully-supervised method on these datasets, demonstrating that it effectively leverages CLIP's open-vocabulary capabilities.

\textbf{Discussion on Spatial Branch.} In ViT, patch features are progressively transformed through successive layers, which critically employ residual connections. This design ensures that features from lower layers are directly added to higher-level ones, thereby retaining a degree of similarity to the features in the final layer and serving as a complement to the final patch features. As demonstrated in \cref{spatial}, the respectable segmentation performance achieved using only the Spatial Branch corroborates this hypothesis. However, the results on the COCO dataset, indicate that feature misalignment can still lead to performance degradation. Thus, there is still scope for refining this approach, for example, by selectively selecting the most suitable lower layers' features.

\textbf{Computational Analysis.} While the main body of this work focused on enhancing the CLIP model's segmentation capabilities, computational efficiency was not a primary design constraint. We now turn to an analysis of the computational cost associated with each component of our method and suggest avenues for its improvement.

The primary computational burden stems from the number of SAM sampling points as shown in \cref{cost}. Reducing uniform sampling from 32$\times$32 to 8$\times$8 substantially improves processing speed with only marginal performance degradation, indicating that many sampled points are repetitive and optimizing sampling strategies is a key direction for acceleration. Another critical avenue for speed improvement lies in the adoption of a more efficient mask generator.

The computational cost of Map Correction is due to the cyclic updating of categories in the mask. Designing a parallel algorithm can further improve its speed.

The computational overhead of Feature Refinement arises from the addition of extra mask class tokens. The computational burden of Value Reconstruction stems from incorporating the DINO model, which can be mitigated by adopting smaller model variants (e.g., ``DINO-S'') to enhance speed and reduce memory footprint.

The clustering algorithm employed in Mask Merging is currently implemented on the CPU, with the potential for future acceleration through the use of GPU.

To enhance the real-world applicability of CorrCLIP, we implement two modifications to mitigate its high computational cost. First, we substitute SAM with more efficient mask generators. Second, we streamline CorrCLIP by removing mask merging and value reconstruction, resulting in a speed increase with an acceptable performance loss. As a result, CorrCLIP now achieves inference speeds comparable to, or even exceeding, those of ProxyCLIP and Trident, while retaining superior performance. Note that EoMT \cite{eomt} and EntitySeg \cite{entityseg} perform better because they are better suited for our method's whole-image segmentation needs than SAM, which is designed for promptable segmentation.

\begin{table}[t]
  \centering
    \resizebox{0.5\textwidth}{!}{
  \begin{tabular}{@{}l|llll@{}}
    \toprule
    Method & FoodSeg103 \cite{foodseg103} & ATLANTIS \cite{atlantis} & CUB-200 \cite{cub-200} & SUIM \cite{suim} \\ 
    \midrule
    OVSeg-L \cite{openvg} & 16.4 & 33.4 & 14.0 & 38.2 \\
    CAT-Seg-L \cite{cat-seg} & 30.5 & 33.6 & 9.2 & 54.0\\ 
    \midrule
    CorrCLIP-L & 
    \textbf{36.5} \textsubscript{\textcolor{blue}{+6.0}} & 
    \textbf{40.1} \textsubscript{\textcolor{blue}{+6.5}} & 
    \textbf{31.3} \textsubscript{\textcolor{blue}{+17.3}} & 
    \textbf{58.0} \textsubscript{\textcolor{blue}{+4.0}}\\ 
    \bottomrule
  \end{tabular}}
  \caption{Comparison with fully-supervised methods on some datasets in MESS.}
  \label{mess}
\end{table}

\begin{table}[t]
  \centering
    \resizebox{0.3\textwidth}{!}{
  \begin{tabular}{@{}l|ccccc@{}}
    \toprule
    Branch & VOC20 & PC59 & Stuff & ADE &City\\
    \midrule
    Main & 88.7 & 46.2 & 30.6 & 25.3 & 48.3\\
    \midrule
    Spatial & 74.5 & 43.8 & 27.7 & 24.1 & 46.3\\
    \bottomrule
  \end{tabular}}
  \caption{Segmentation performance achieved using only the Spatial Branch.}
  \label{spatial}
\end{table}

\begin{table*}[t]
  \centering
  \resizebox{0.75\textwidth}{!}{
  \begin{tabular}{@{}l|ccc|c@{}}
    \toprule
    {Method}  & Time(ms/image) $\downarrow$ & Memory(MB) $\downarrow$ & Parameter(M) $\downarrow$ & Performance(mIoU) $\uparrow$ \\ 
    \midrule
    {ClearCLIP \cite{lan2024clearclip}} & 19 & 718 & 150 & 38.1 \\
    {ProxyCLIP \cite{proxyclip}} & 69 & 1936 & 235 & 42.3 \\
    {Trident \cite{trident}} & 81 & 2487 & 364 & 45.8 \\
    \midrule
    \rowcolor{gray!8}
    \multicolumn{5}{c}{Sampled Points: 8$\times$8}  \\
    + Sope Reconsturction& 116 & 2674 & 373 & 45.3 \\
    + Map Correction & 119 & 2674 & 373 & 47.5 \\
    + Feature Refinement & 132 & 2674 & 373 & 49.7 \\
    + Value Reconstruction & 170 & 2997 & 458 & 50.2 \\
    + Mask Merging & 177 & 2997 & 458 & 50.4 \\
    \midrule
    \rowcolor{gray!8}
    \multicolumn{5}{c}{Sampled Points: 32$\times$32} \\
    + Sope Reconstruction & 1111 & 2902 & 373 & 46.8 \\
    + Map Correction & 1115 & 2902 & 373 & 49.2 \\
    + Feature Refinement & 1129 & 2902 & 373 & 50.5 \\
    + Value Reconstruction & 1168 & 3208 & 458 & 50.8 \\
    + Mask Merging & 1258 & 3208 & 458 & 51.0 \\
    \midrule
    \rowcolor{gray!8}
    \multicolumn{5}{c}{Faster Version} \\
    CorrCLIP\tiny{\textcolor{blue}{~M2F}} & 81 & 1765 & 366 & 49.2 \\
    CorrCLIP\tiny{\textcolor{blue}{~EoMT}} & \textbf{56} & 1818 & 466 & \textbf{51.6} \\
    CorrCLIP\tiny{\textcolor{blue}{~EntitySeg}} & 78 & \textbf{1100} & \textbf{197} & 51.1 \\[-2pt]
    \bottomrule
  \end{tabular}
  }
  \caption{Computational costs on RTX 4090 with FP16. Performance is average mIoU across eight benchmarks. The blue subscript indicates the mask generator. M2F (Mask2Former \cite{mask2former}) uses a Swin-L backbone, and EoMT uses ViT-L; both are pretrained on COCO Panoptic. EntitySeg uses a Swin-T backbone.}
  \label{cost}
\end{table*}

\begin{figure*}[ht]
    \setlength{\belowcaptionskip}{0pt} 
  \centering
   \includegraphics[width=0.8\linewidth]{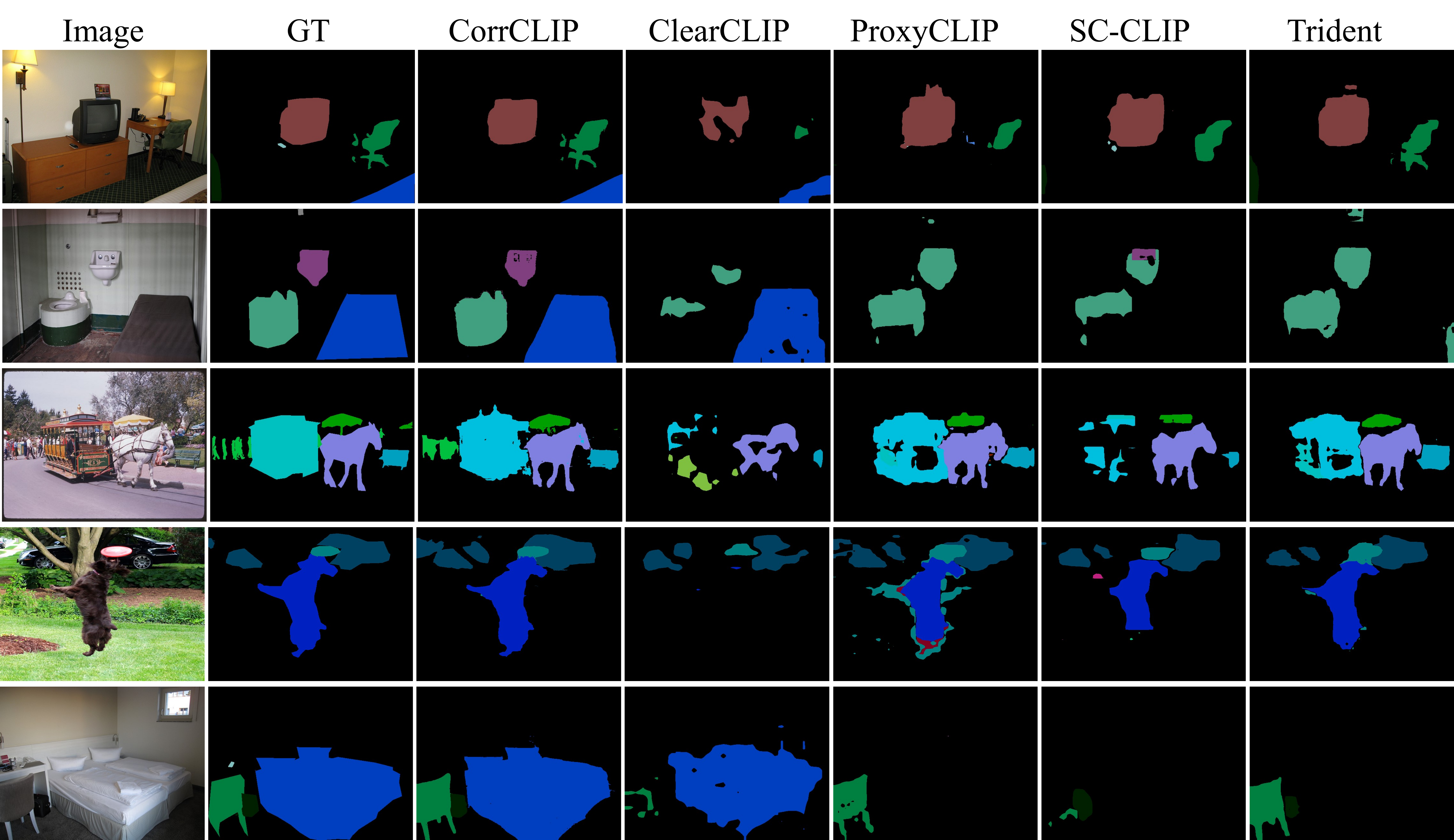}
   \caption{More qualitative comparison of segmentation maps between our method, CorrCLIP, and the other four methods on Object.}
   \label{object}
\end{figure*}

\begin{figure*}
    \setlength{\belowcaptionskip}{0pt} 
  \centering
   \includegraphics[width=0.8\linewidth]{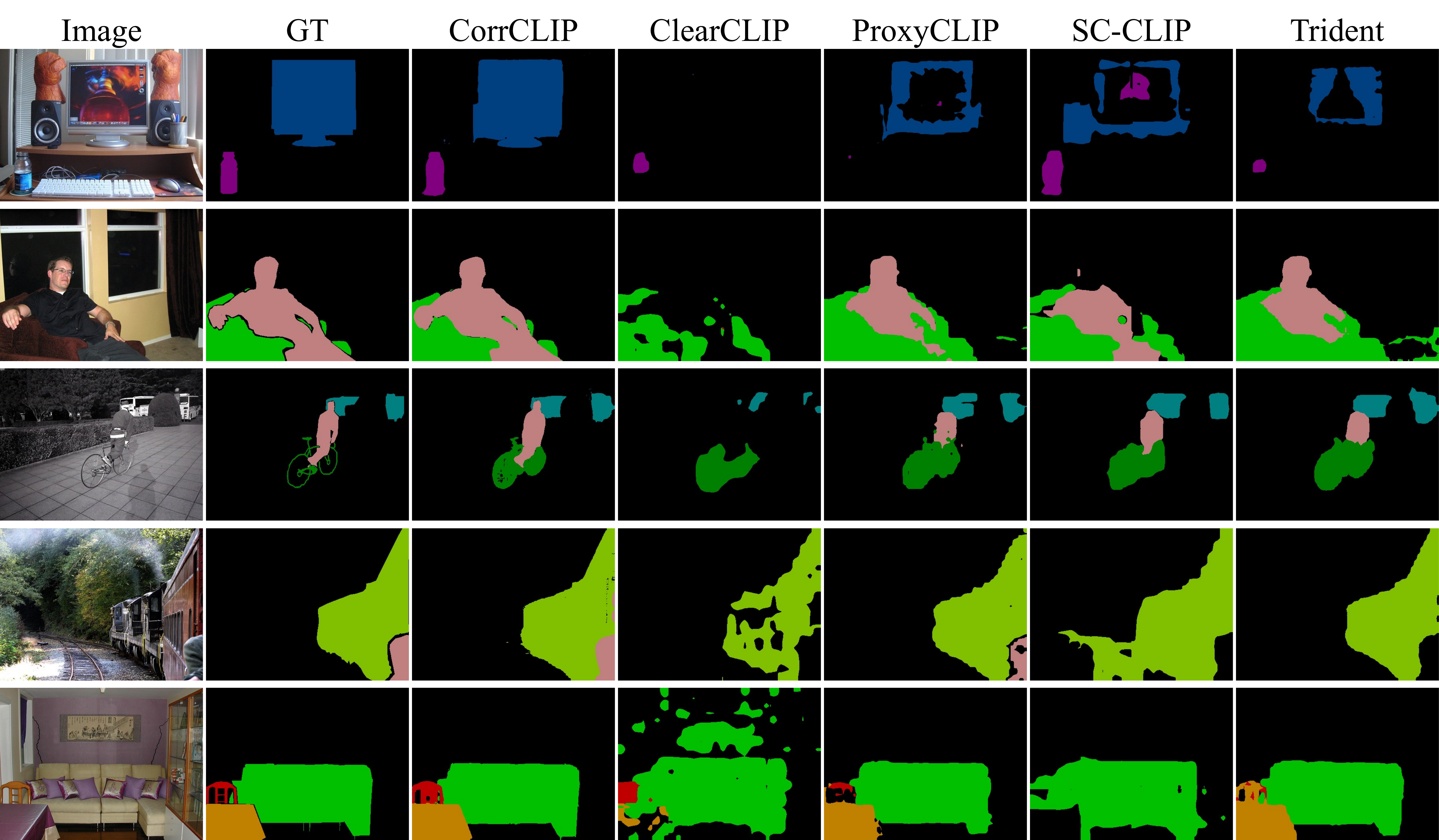}
   \caption{More qualitative comparison of segmentation maps between our method, CorrCLIP, and the other four methods on VOC21.}
   \label{voc21}
\end{figure*}

\begin{figure*}
    \setlength{\belowcaptionskip}{0pt} 
  \centering
   \includegraphics[width=0.8\linewidth]{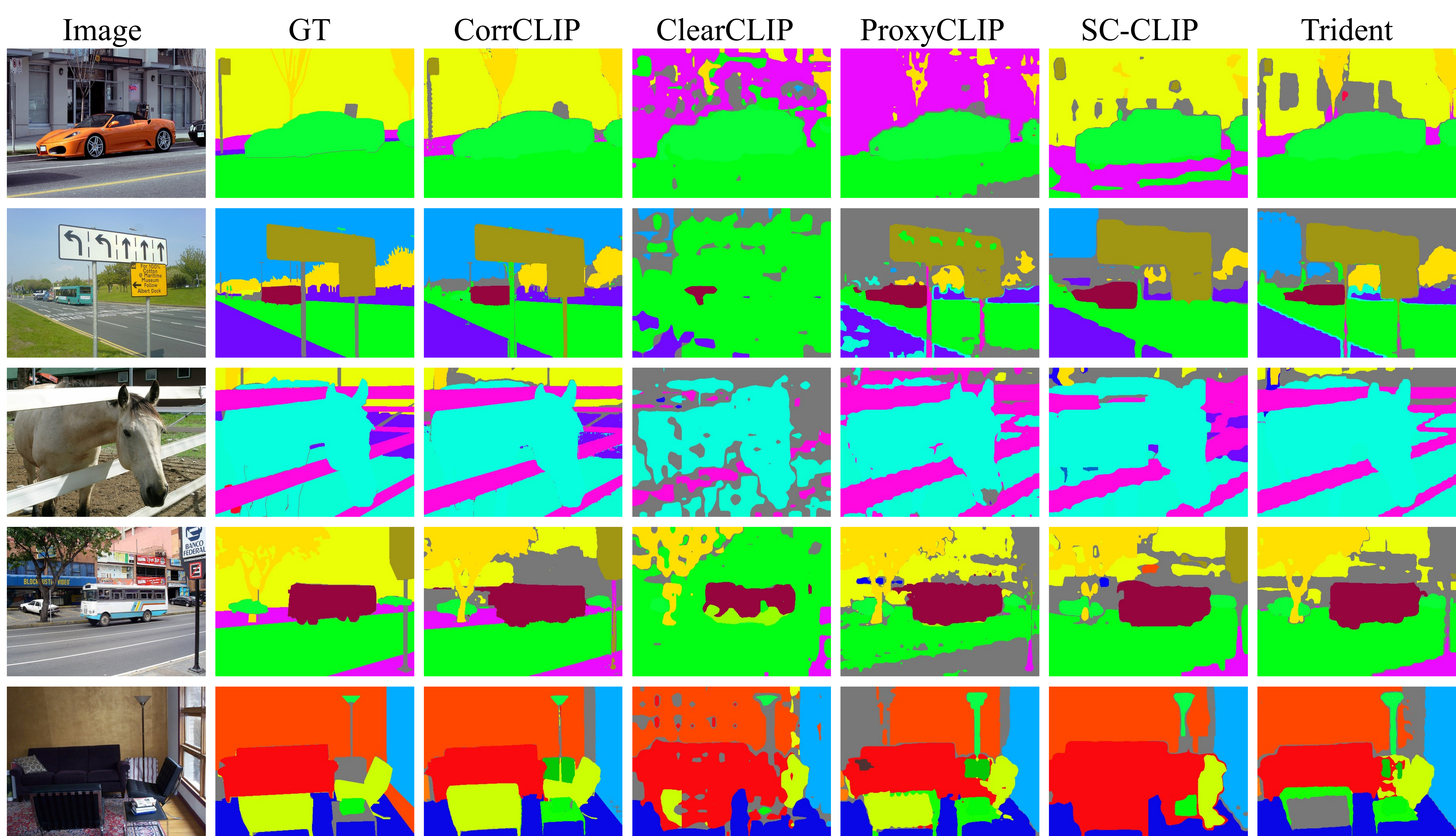}
   \caption{More qualitative comparison of segmentation maps between our method, CorrCLIP, and the other four methods on PC60.}
   \label{pc60}
\end{figure*}

\begin{figure*}
    \setlength{\belowcaptionskip}{-10pt} 
  \centering
   \includegraphics[width=0.8\linewidth]{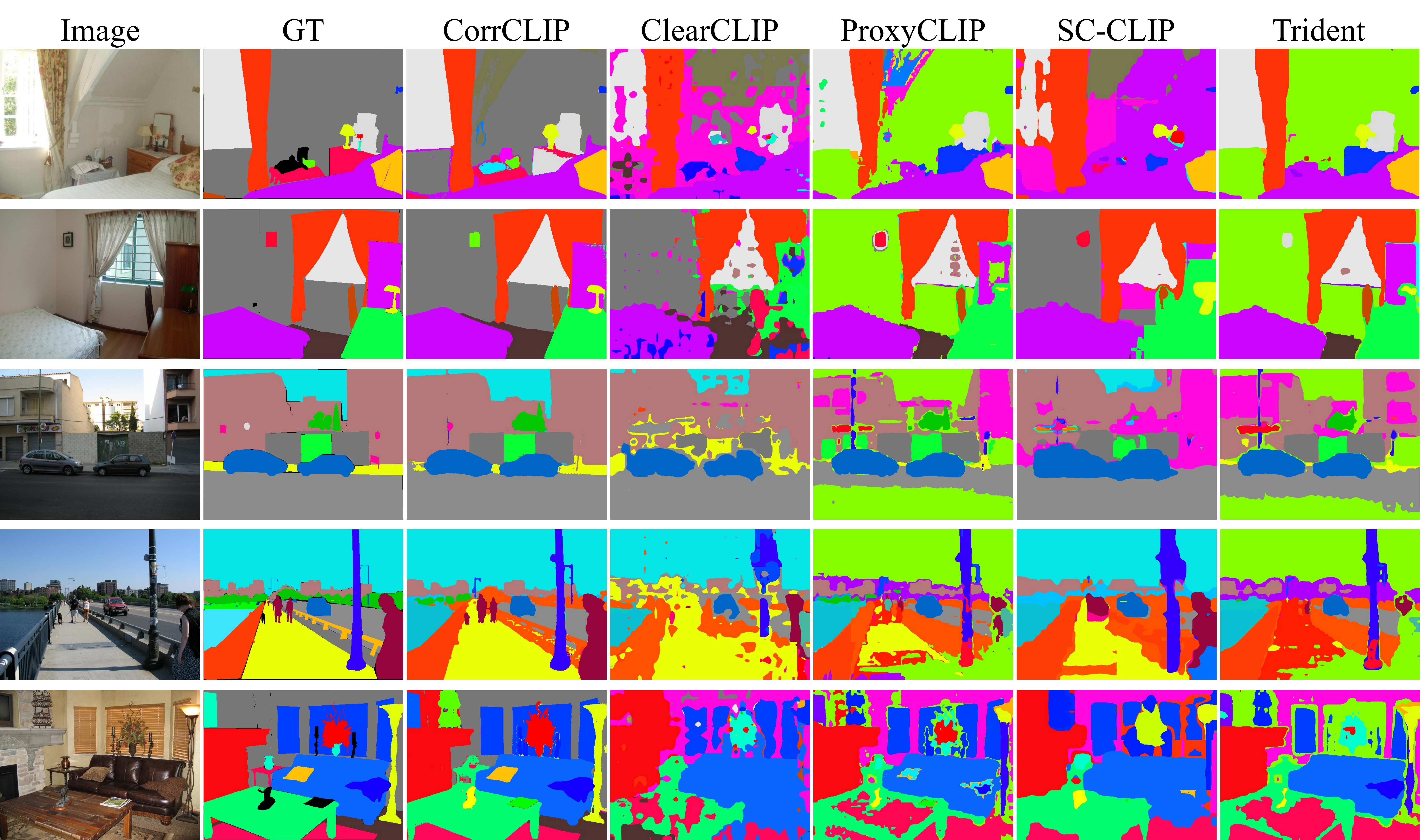}
   \caption{More qualitative comparison of segmentation maps between our method, CorrCLIP, and the other four methods on ADE.}
   \label{ade}
\end{figure*}

\begin{figure*}
    \setlength{\belowcaptionskip}{-10pt} 
  \centering
   \includegraphics[width=0.8\linewidth]{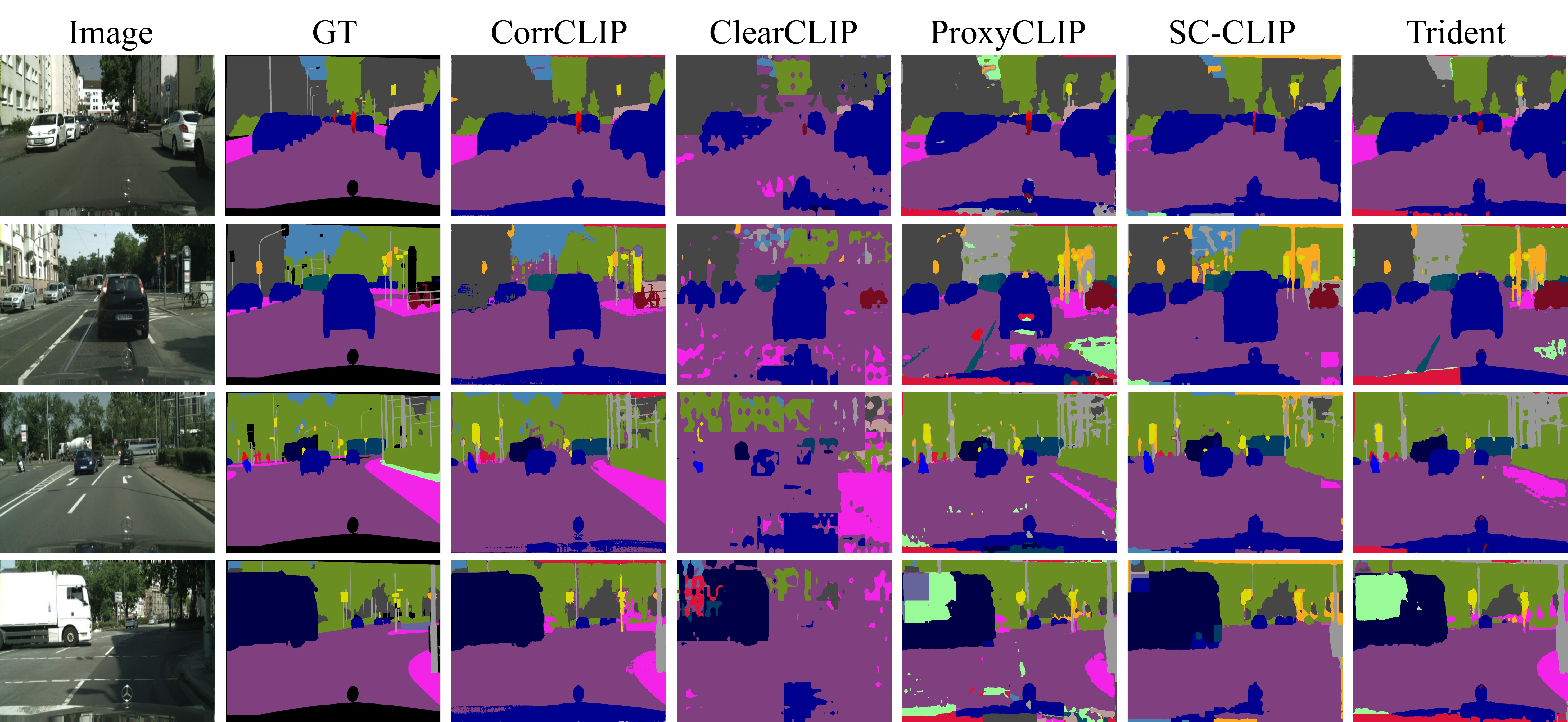}
   \caption{More qualitative comparison of segmentation maps between our method, CorrCLIP, and the other four methods on City.}
   \label{city}
\end{figure*}

\textbf{More Qualitative Comparisons.} We provide more qualitative comparisons between our CorrCLIP and existing methods, including ClearCLIP \cite{lan2024clearclip}, ProxyCLIP \cite{proxyclip}, SC-CLIP \cite{sc-clip}, and Trident \cite{trident} in \cref{ade,object,city,voc21,pc60}. Through explicit interaction scope restriction, our approach successfully corrects misclassifications that persist in other methods. Additionally, CorrCLIP exhibits three notable advantages: superior object continuity preservation, effective noise suppression, and enhanced robustness in challenging scenarios where alternative methods often exhibit fragmentation or semantic inconsistency. These qualitative results corroborate our method's capability to establish more reliable visual-semantic correspondences.

\end{document}